# Deep Reinforcement Learning with Macro-Actions

**Ishan P. Durugkar**
University of Massachusetts,
Amherst, MA 01002
idurugkar@cs.umass.edu

**Clemens Rosenbaum**
University of Massachusetts
Amherst, MA 01002
cgbr@cs.umass.edu

**Stefan Dernbach**
University of Massachusetts
Amherst, MA 01002
dernbach@cs.umass.edu

**Sridhar Mahadevan**
University of Massachusetts
Amherst, MA 01002
mahadeva@cs.umass.edu

## Abstract

Deep reinforcement learning has been shown to be a powerful framework for learning policies from complex high-dimensional sensory inputs to actions in complex tasks, such as the Atari domain. In this paper, we explore output representation modeling in the form of temporal abstraction to improve convergence and reliability of deep reinforcement learning approaches. We concentrate on macro-actions, and evaluate these on different Atari 2600 games, where we show that they yield significant improvements in learning speed. Additionally, we show that they can even achieve better scores than DQN. We offer analysis and explanation for both convergence and final results, revealing a problem deep RL approaches have with sparse reward signals.

## 1 Introduction and Related Work

Since the groundbreaking results shown by Deep Q-Learning in [10, 11] for learning to play games from the Atari Learning Environment [4], there has been extensive research on deep reinforcement learning. Deep Q-learning in particular seeks to approximate the Q-values [21] using deep networks, such as deep convolutional neural networks [10]. There has also been work on modifying the loss used to train the network using more consistent operators [5], or ensuring better target estimation [19]. Generally, research on improving the learner's performance can broadly be classified along the following directions: (i) improvements to the deep network [20, 6, 13] (ii) improvements to the reinforcement learning algorithm [5, 19] and (iii) implementation enhancements, such as performing better gradient updates [9, 12] and prioritizing experience replay to maximize learning [15]. In this paper, we focus on the second category of improvements.

In reinforcement learning (RL), temporal abstraction [14] denotes the use of hierarchical multi-step actions that the agent can take in the possible addition to the available primitive actions. In terms of Markov decision processes (MDPs), a well-studied framework for analyzing reinforcement learning algorithms, temporal abstraction allows executing multi-step actions, represented by entire paths in the MDP. An MDP that has been extended to allow modeling such paths is sometimes referred to as a semi-MDP (SMDP) [16]. Since one cannot consider all possible abstractions–the number of such abstractions grows exponentially with their length–learning useful (or meaningful) abstractions has long been an important problem in reinforcement learning [3].

Some recent work on hierarchical models [8, 1] aim to achieve hierarchical behavior using closed loop control, or options [17]. The drawback of these methods is that they require external supervision, either by explicitly specifying the mapping from states to options [1] or by specifying possible sub-goals [8].

Another possible set of abstractions are state-independent, deterministic sequences of actions, called macro-actions or macros [7]. A macro can be chosen as if it was an atomic action, and the agent will then (deterministically) follow the sequence of actions predefined by the macro. Consequently, intermittent states are visited, but are non-actionable by the agent. However, any reward following from the visited state is still accumulated.

We explore the effect temporal abstraction has on deep RL by testing the effectiveness of using macros in addition to basic actions. Specifically we extend the Deep Q-Network [11] to use macro actions. It is important to note that even though our evaluation is based on the the original DQN architecture, our technique does not affect the remaining architecture, and can be paired with other improvements as mentioned above. We evaluate the effectiveness of macros, and contrast different approaches to construct them. We then illustrate how a reinforcement learner with access to macros interacts with deep networks, showing how macros can improve reasoning on deep network input representations.

## 2 Hierarchical Reinforcement Learning

Hierarchical RL in general can be motivated along many lines. Among those are a (potential) solution to the "curse of dimensionality"–the problem that solving a problem becomes exponentially harder with its size. Here, a hierarchical RL agent can offer an elegant divide-and-conquer approach by separating the problem into smaller sub-problems. Further advantages are architectural; a hierarchical learner can add to parametrizable actions, to continuous action- and state spaces and to generalizability of policies. We will discuss the benefits particular to macros in section five.

An important distinction in hierarchical RL is between open loop and closed loop models. Closed loop models can be loosely related to a functional perspective: "hit your opponent" or "go find the key, and open that door". In this case, the path is not described by its internal structure, but by it's terminal states instead. In terms of temporal abstraction, this problem is most closely modeled by options [17], i.e. stochastic policies defined over a subset of the (S)MDP. However, hierarchies of this kind are difficult to learn, since they require both a composition of atomic actions and a decomposition of a larger goal into smaller subgoals (an illustration of manual goal-decomposition is given by [8]).

Open loop models can be related to a more structural perspective: "move five steps left, three up, and punch" would replace "hit the opponent". This approach is both more and less useful than the first. Once successfully learned, closed loop strategies can be much more useful than open loop strategies, since they are more general.On the other hand, viable sub-goals are hard to define in a general way for an agent to recognize on its own. Generally some sort of supervision is provided to define these sub-goals. In this work, we concentrate on open loop models in the form of macros exclusively.

| | Notation and Definitions |
|---|---|
| $\pi$ | Policy mapping from states to a probability distribution over actions. |
| $\mathcal{S}, \mathcal{A}, s, a$ | Sets of states and actions and an individual state and action, respectively. |
| $R(s,a), r_t$ | Reward of taking action $a$ in state $s$ and a specific reward received at time $t$, respectively |
| $P(s'\|s,a)$ | Transition probability from state $s$ to $s'$ when taking action $a$ |
| $V(s)$ | Value of state s. |
| $\gamma$ | Discount factor used to determine the effect of future reward. |
| $Q(s,a)$ | State-action (or $Q$) function, determining the value of taking action $a$ in state $s$. |
| $\mathcal{M}, \|\mathcal{M}\|, \ell, m_i$ | Macro-specific variables (the set of macros, the size of the set, the length of a macro, and the $i^{th}$ macro) |

### 2.1 Background

Hierarchical RL is generaly modeled on semi-MDPs (SMDP). While in an MDP, each state transition is assumed to occur at uniform time intervals (encoding a one-step path), an SMDP is a generalization of an MDP that adds a random variable encoding the time difference in between successive states (thereby possibly encoding paths of arbitrary length). This random variable can be subject to certain additional constraints–it can be real-valued, discrete or subject to any other restrictions defined by the domain. In our case, the domain, the ALE [4], is a discrete time step system that can be interacted with every 1/15 s.[1]

An important consequence of changing an MDP to an SMDP is that the fixed point of the value function for a given policy changes:

$$\text{from} \quad V_\pi(s) = R(s, a_\pi) + \gamma \sum_{s'} P(s'|s, a_\pi) V_\pi(s') \quad (1)$$

$$\text{to} \quad V_\pi(s) = R(s, a_\pi) + \sum_{s', \tau} \gamma^\tau P(s', \tau|s, a_\pi) V_\pi(s') \quad (2)$$

where $\tau$ represents the possible duration of time taken by a temporally extended action. Consequently, the Q-learning update underlying many deep reinforcement learning algorithms changes

$$\text{from} \quad Q_{k+1}(s, a) = Q_k(s, a) + \alpha_k \left[ r_k + \gamma \max_a Q_k(s', a') \right] \quad (3)$$

$$\text{to} \quad Q_{k+1}(s, a) = Q_k(s, a) + \alpha_k \big[ r_{t+1} + \gamma r_{t+2} + ...$$
$$+ \gamma^{\tau-1} r_{t+\tau} + \gamma^\tau \max_a Q_k(s', a') \big] \quad (4)$$

where $r_{t+1}, r_{t+1}, ..., r_{t+\tau}$ are the rewards received in-between actionable states.

### 2.2 Macros

Macro-actions or macros are predefined sequences of actions that will be taken deterministically [7]. Using $\mathcal{M}$ to refer to the list of all macros, $m_i$ to refer to the $i^{th}$ macro within this list and $\ell_i$ to the length of the $i^{th}$ macro, we can then define a macro as:

$$m_i = \langle a_{i,1}, ..., a_{i,\ell_i} \rangle$$

Here, $a_{i,x}$ can be any of the actions available to the agent, generally including other macros although we do not explore this direction in this paper. Additionally, we use $|\mathcal{M}|$ to refer to the number of macros.

As an example consider the Atari game "Boxing". In our experiments, we find that sequences of the kind "up, up, punch, down, down", representing moving in for a punch and pulling out are fixed sequences that can be particularly useful.

Since we restrict our discussion to the Atari domain for simplicity, and this domain is deterministic, we can make the following observations:

1. The states $\mathcal{S}_i = \langle s_s, s_{s+1}, ..., s_{s+\ell_i} \rangle$ visited by executing macro $m_i$ in state $s_s$, can be determined by the sequence of actions in the macro.

2. Consequently, the terminal state $s_{s+\ell_i}$ is the last state visited, and is determined by the sequence of actions.

3. The cumulative reward collected is the sum of the rewards of taking the actions defined in the macro, beginning from the original state: $\mathcal{R}_{s_s, m_i} = \sum_{k=1}^{\ell_i} r(a_{i,k}, s_{s+k})$

While we will discuss the most prominent benefits macros offer later, we want to point out their cost here: Adding macros increases the size of the action space, which worsens the effect of high dimensionality. Using macros also means the agent skips decision making at certain states, which could lead to a sub-optimal policy.

## 3 Learning Macros

In this section, we highlight the different challenges in choosing useful macros, and our modifications to DQN to address them. We also discuss different approaches to vary the shape of the macros, i.e. the sequence of actions contained. While the aim is to learn macros, we include another approach that isu meant as a baseline for later analysis.

---

[1] The internal clock of the ALE runs at 60 frames per second, and each action is active for 4 frames, this extension through time could already be considered a case of temporal abstraction. However, since our learning agent considers the ALE to be a "black box", we consider the 1/15 s interactions to be atomic.

**Algorithm 1** Changes to the DQN incorporating macros
---
**repeat**
  · Follow the DQN algorithm.
  **if** epoch = in $\mathcal{K}$ **then**
    · Compute the new list of macros by the designated policy; if longer than $|M|$, cut off the last macros. If shorter, fill the list with empty macros and disable them as actions.
    · Add the macros as actions to the DQN algorithm.
    · Reset the observed action sequences.
  **end if**
**until** epoch = maxNoEpoch
---

When deciding on open loop sequences of actions, the major parameters to attend to are the length of the macros, the number of macros to use and the exact sequence of actions that make up these macros. Shorter and lower number of macros are not much different from atomic actions, while longer macros might delay the agent's responses. A higher number of macros will increase the number of actions the agent has to learn over. Once these parameters are decided, we turn to the problem of what actions these macros should be made up of.

We test our hypothesis on the Atari [4] domain using an agent similar to the DQN [11]. The input to the agent is the screen image from the ALE, taken in grayscale and resized to an $(84 \times 84)$ frame. As in the original DQN, we take 4 consecutive frames to denote one state. These frames are then analyzed by a deep neural net structurally the same as the original DQN, 3 convolution layers, one fully connected layer and one output layer, with corresponding layer sizes except for the output layer. The updates are computed using the variation to RMSProp [18] used by [11]. The only structural difference between the DQN of [11] and our model is that the output layer is expanded to include the macros.

To account for the temporal aspect of actions that last longer, when saving transitions in the replay memory we also include $\tau$, or how much time each transition took, and accumulate rewards as $R_t = r_{t+1}, r_{t+2}, ..., r_{t+\tau}$. This $\tau$ is then used while calculating the Bellman error. To incorporate the macros, we modify the network by expanding the output layer. The size of the output layer is taken to be $|\mathcal{A}| + |\mathcal{M}|$. We initialize $|\mathcal{M}| = |\mathcal{A}|$.

If we learn new macros, and the set of macros is smaller than the size of the output layer, we ignore extraneous outputs while choosing the next macro to execute, or while choosing the best Q-value at the next state when calculating the target for the Bellman error. This does not affect the backpropagation of error at any step since the error is only propagated via the output associated with the action taken during that transition. This allows us to change the size of the macros online. We replace macros at certain epochs with increasing gaps in between these replacements, to give the agent more time to exploit this new action-space.

Since modifying macro assignment with respect to the output layer might mean that an output no longer takes the same sequence of actions it did earlier, the Value for that output needs to be learnt again, and the weights need to be adjusted accordingly. We do this by increasing the exploration factor to $\epsilon = 0.5$ and allowing it to decay in accordance with DQN, rather than reinitializing the weights and having the agent learn them from scratch again.

We exclude the possibility of macros containing macros, but consider first-order abstractions (i.e. macros built on atomic actions) exclusively, which we append to the list of atomic actions.

### 3.1 Repetition of Actions

In this method, each macro is composed of an atomic action repeated a specified number of times, with a macro corresponding to each action. In effect, this amounts to making the actions more granular, or taking larger steps with the same action. These macros take advantage of the fact that the time step of each action in the Atari domain may be smaller than is actually required to take optimal decisions.

$$|\mathcal{M}| : \text{chosen as a parameter} \qquad \ell_i : \text{chosen as a parameter} \qquad a_{i,k} : a_i, \ a_i \in \mathcal{A}$$

---

**Algorithm 2** Selecting macros by execution frequency at the end of an epoch in $k$

---

    Initialize an object to count action sequence occurrences, $\mathcal{O}$
    **for** $i$ `from 0 to` $|\mathcal{A}| - \ell$ **do**
        Increment the counter in $\mathcal{O}$ of the sequence $\mathcal{A}[i : i + \ell]$ by 1
    **end for**
    Rank all sequences by their occurrence in $\mathcal{O}$
    Initialize the set of macros $\mathcal{M}$ to contain the highest-occurring sequence
    **repeat**
        Take the next highest occurring sequence from $\mathcal{O}$, $s_*$
        **if** the longest common subsequence between $s_*$ and all sequences in $\mathcal{M}$ is less than $\omega\ell$ **then**
            Append $s_*$ to $\mathcal{M}$
        **end if**
    **until** $|\mathcal{M}|$ is reached or all sequences in $\mathcal{O}$ are exhausted

---

### 3.2 Frequency

Here, the macros are chosen by picking sub-sequences of actions of specified length that are repeated the most often during the trajectories since the last update of the list of macros, as long as they are sufficiently different. Difference is measured by the length of the longest common subsequence (lcs) in between two macros. If the lcs between a macro candidate and any other macro is longer than a percentage $\omega$ of the total length of the macros $\ell$, then the candidate is dropped. See Algorithm 2 for a more detailed explanation.

$|\mathcal{M}|$ : chosen as a parameter          $\ell_i$ : chosen as a parameter          $\omega$ : chosen as a parameter

$\ldots a_{i,0}, a_{i+1,1}, \ldots, a_{i+n,\ell} \ldots$ is the $i^{th}$ most repeated sequence by the agent if sufficiently different from the $[0, i-1]^{th}$ most repeated sequences. Sequences repeated most often by the agent can be assumed to be useful to the agent given the current policy.

## 4 Experimental Results

We evaluate this technique on the Arcade Learning Environment [4], a reinforcement learning interface to Atari 2600 games.

For our evaluation, we train agents following Algorithm 1. We have trained the agents on 8 games. To showcase the speed of convergence using macros, we show the performance of the agent after 50 million frames. We compare this behavior to our DQN implementation based on [11] trained for 100 million frames, i.e. twice the training period. Testing is done after each epoch (1 million frames) for 200,000 frames to evaluate the agent's performance. To showcase stability and performance, the performance scores are averaged over 5 epochs, and variance is shown over these as well.

To learn macros based on frequency, we allow as much as $\omega = 80\%$ overlap and we update these macros with newer learned ones at larger and larger intervals (in epochs 6,13,25 and 50). This is because as an agent starts converging on a policy, changing the macros assigned to the network outputs tends to degrade performance before the agent can relearn the Q-values for the new macros. Larger intervals let the agent acclimatize to the new action-space, and converge on a policy. This can be seen from Figure 1b and 1d, where the red dashed lines indicate epochs where the agent learns new macros. At these points, the performance of the agent tends to go down, and the variance increases, until the macros converge as well .

There is some variance in our baseline as compared to [11]. One of the primary reasons is that we are showing the mean of average scores per episode over 5 epochs, whereas in the original paper the result shown was average over the best scores achieved. The results are shown in Table 1.

The main advantage we gain from macros is improved convergence times. This is evident from the results shown after training for 50 epochs. The plots over 10 trials of pong and Enduro also show the convergence and stability for both action repetition and frequency analysis based macros.

| Game | Baseline | Repetition-3 | Repetition-5 | Frequency-3 | Frequency-5 |
|---|---|---|---|---|---|
| Breakout | **259.71**(63.64) | 211.43(46.2) | 197.2(72.3) | 228.7(**35.26**) | 227.43(52.8) |
| Enduro | 457.65(125.24) | 461.31(70.08) | 448.7(88.4) | 535.36(**62.38**) | **605.28**(111.79) |
| Pong | 20.7(0.45) | **21.0**(**0.0**) | 17.94(0.35) | **21.0**(**0.0**) | **21.0**(**0.0**) |
| Boxing | 76.66(8.44) | 82.1(7.27) | 79.5(8.3) | **91.23**(**5.68**) | 89.52(7.21) |
| Ms. Pacman | 2204.79(228.84) | **2426.16**(359.0) | 2025.5(186.26) | 2152.36(**178.2**) | 2357.23(225.12) |
| Space Invaders | 922.15(143.76) | 920.36(94.82) | 883.76(101.2) | **933.3**(**87.5**) | 921.43(92.3) |
| Qbert | 5261.12(1604.27) | 5146(784.79) | 3712.65(981.26) | **5303.13**(**542.88**) | 5212.47(723.91) |

Table 1: Mean Score for different games with length 3 and 5 macros along with deviation over multiple epochs. The scores for macros are calculated at 50 epochs of training to showcase speed of convergence. Higher mean scores and lower deviations are highlighted in bold.

## 5  Analysis

There are three different measures of performance we can evaluate macros on: scores achieved, the (arguably) most important measure; speed of convergence; and variance of scores over some epochs. To explain the respective performance of macro-actions, we offer and evaluate three hypotheses. The first is the impact macros have on exploration, the second is the impact they have on error-propagation and the third is the approximation precision of the deep representation, with its possible challenge to greedy algorithms (like the Q-learning algorithm used here).

From the results, it can be seen that an agent with access to useful macros performs better than an agent with access only to atomic actions. We analyze this behavior in the context of both speed of convergence and the score achieved.

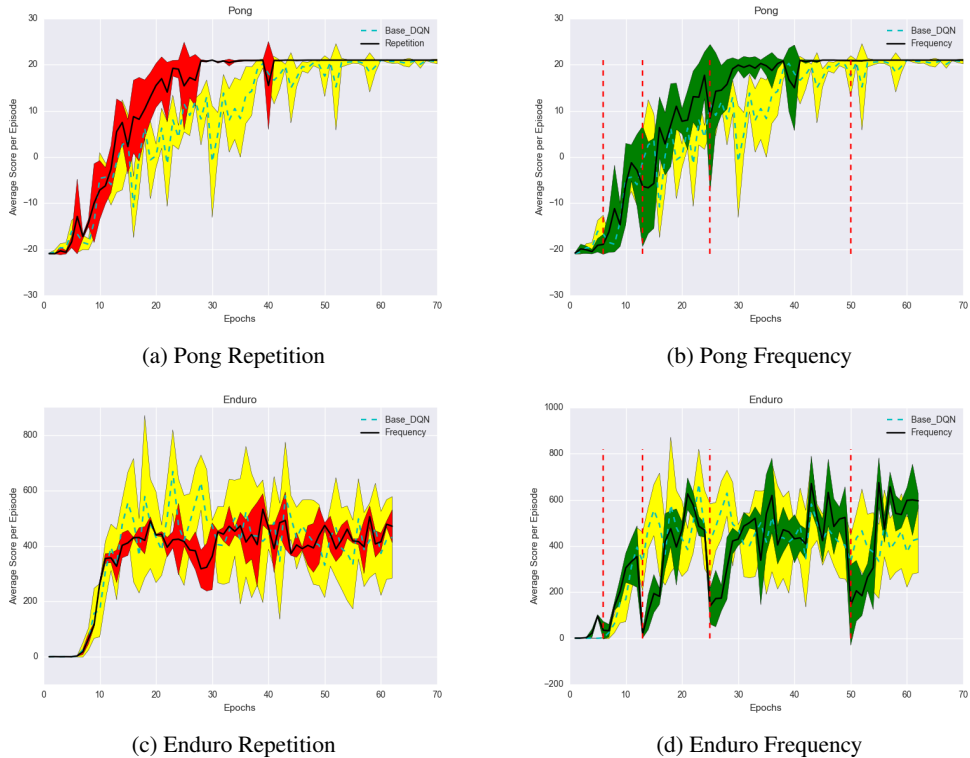

(a) Pong Repetition  (b) Pong Frequency

(c) Enduro Repetition  (d) Enduro Frequency

Figure 1: Convergence Behavior on Pong (1a and 1b) and Enduro (1c and 1d). 1a and 1b showcase the acceleration in convergence while 1c and 1d are indicative of the low variance of the policy learnt. The red dashed lines denote epochs at which new macros are learnt, explaining the drop in performance for the learned macros. The blue dashed line is the mean over 10 trials of the Base DQN. The solid lines denote action space expanded using macros. The colored area shows variance.

### 5.1 Explanatory Hypotheses

As mentioned previously, there are two major established benefits to macros which we will discuss now. The first is related to the exploration-exploitation dilemma in reinforcement learning, i.e. the dilemma in between exploiting already gained knowledge and trying something new. Exploratory moves using macros offer larger exploration in a directed manner. Considering that the games in the Atari domain are very complex and consist of millions of states, more efficient exploration should hence offer convergence benefits.

The second potential benefit macros offer is the changed propagation of rewards throughout the SMDP. This change stems from macros receiving the cumulative reward along the states visited, thereby propagating rewards from states up to $\ell$ steps farther ahead in one iteration; in the MDP setting, this takes up to $\ell$ times more iterations, since the reward is only propagated one state in each update. This implies that the Bellman error is equally propagated much faster, allowing for fewer iterations until convergence.

### 5.2 Convergence

Considering that rate of convergence is macros' most reliable benefit, it does not surprise that the same holds for macros used in the Atari domain. However, there are two potential reasons, first the exploratory bias and second the faster propagation of values. While it is near impossible to perfectly quantify the impact each of these two reasons has, we can still draw some conclusions on how the different macros behave. To do so, we contrasted the performance of the well-defined macros introduced previously with macros randomly populated. These do offer better exploration, since they will force the agent to explore states and regions further away. However, they do not offer substantially better convergence than the pure DQN approach and perform significantly worse than both manual macros and learned macros. Since they do not necessarily offer a higher "spread" over the MDP the propagation of rewards may be less, this suggests that faster convergence may be more heavily influenced by the propagation of values than by the macros' exploratory bias.

Additionally, we can infer that learning macros does indeed learn useful macros, since it offers convergence behavior similar to manual macros, considerably outperforming the basic DQN.

### 5.3 Variance

As can be seen in the results, the final variance of the scores achieved is consistently lower when using macros (for Qbert in particular, the variance is dramatically smaller, while achieving higher final scores). We believe that this results from a combination of better exploration and better propagation of rewards. Using atomic actions only, the agent may not consistently explore areas with higher Q-values, which may lead the learner for finite training time to learn different policies. Higher exploration, and better propagation of the rewards, will make this less likely, since the learner will explore more distant states, and will have states with Q-values differing to choose from. Conclusively, macros lead to a policy where the agent has high confidence for the best action, leading to more stable policies.

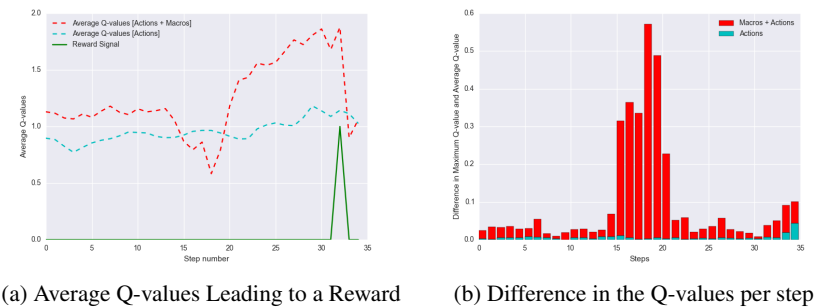

(a) Average Q-values Leading to a Reward    (b) Difference in the Q-values per step

Figure 2: Comparison of changing Q-values and the Difference in Q-values for states leading up to a reward in the game of Pong. Values in red are for an agent using Macros + Actions and values in blue are an agent using just Actions. The bar graph is stacked to show comparison

### 5.4 Scores Achieved

**General observations**    An agent with access to macros does better on 6 out of the 7 games tested on than an agent with access to only atomic actions. The only game in which the behavior is at or below par with atomic actions is Breakout.

This is a surprising result, since coarse control in general should not yield better results. After all, an agent using atomic actions can always choose the same actions as enforced by a macro. We believe that this follows from a third benefit macros offer when combined with a deep network, larger resistance to error when approximating the Q-values of a state. Most games in the Atari domain offer sparse rewards (reward feedback is available after hundreds or in some cases even thousands of frames). Considering discounting and the ability to correct mistakes in intermediate states, this entails that states farther away from a potential reward will have Q-values within a very small $\delta$ around their average. This line of thought becomes obvious when looking at the plots in Figure 2. Figure 2b shows that the "confidence" of the Q-learner, in terms of difference of the Q-values the greedy `argmax` chooses from, is orders of magnitude higher when using macros. Figure 2a shows the absolute values for reference. Please note that the behavior shown is for a policy learned in 100 million steps in the DQN, but only in 50 million for the macro case.

Further, the Q-value estimates of the network will contain some inherent approximation error. This error combined with the the underlying actual Q-values being similar means that the greedy `argmax` of a Q-learner will in some cases not be able to pick out the best action. Macros, with steps that change the target states more substantially, will therefore offer Q-value approximations that are more spread out, superseding the approximation error. This leads to more reliable decision making for the Q-learner, which eventually results in better policies learned. This relates macros to advantage learning, since they, too, increase what is known as the action gap [5, 2].

Additionally, this benefit is not exclusive to the reinforcement learning part of the deep RL architecture, but affects the neural network that is used to learn a useful representation of the state of the game as well. The network has to learn to differentiate between and meaningfully represent the changes in the state every time an action is taken. When the agent takes temporally extended actions, the network is given the opportunity to learn changes that affect the state in a more noticeable manner. This means that the network can more quickly encode these changes, and more of the training bandwidth is expended in tuning the agent policy.

**Approach-specific observations**    Considering their respective results, it is interesting how well the simple repetition of actions performs. Being considerably worse in Breakout is most likely a consequence of Breakout's requirement for fine-grade control. Temporal abstraction as provided by extended sequences of singular actions interferes with this process. However, it performs best for Ms Pacman. The argument to be made here is that Ms Pacman repeats actions by itself (i.e. once "up" is pressed, the agent will move up until it hits a wall). Any fine interference, either by macros of multiple actions, or by atomic actions for exploration, will interrupt this process.

It is equally interesting how little influence learning the macros by evaluating their frequency has. Intuitively, this would lead to macros that model specific strategies. However, this only holds true for Boxing, where the agent indeed learns strategies that resemble patterns such as "go in, punch, and leave".

## 6   Conclusion and Future Work

We have shown how macro-actions can improve the convergence times and scores of a DQN agent on the Atari domain. Not only do macros improve the efficiency of the agent's training, but also help the agent achieve better scores on six out of the seven games we tested.

We have analyzed the reasons why macros can improve the policies learned by a deep reinforcement learning agent. Useful macros are likely to accelerate the learning of the agent and encourage discovery of optimum policies. Better macro discovery techniques would be a viable next step in this direction. Considering the different aspects of learning macros, we should ideally develop a technique that can not only learn the actions to be taken, but can learn their optimal length and quantity as well.

As open loop policies, macros take actions in a specified sequence without looking at the underlying state once it is initiated. While this means that the agent does not need to take decisions as frequently, and learns quicker, macros can be seen as a architecture too rigid for many problems. It would hence be interesting to develop a similar way of applying closed loop models, such as options, to a deep reinforcement learning agent and to compare its performance on the Atari domain.

# References

[1] Kai Arulkumaran, Nat Dilokthanakul, Murray Shanahan, and Anil Anthony Bharath. Classifying options for deep reinforcement learning. *arXiv preprint arXiv:1604.08153*, 2016.

[2] Leemon C Baird III. Reinforcement learning in continuous time: Advantage updating. In *Neural Networks, 1994. IEEE World Congress on Computational Intelligence., 1994 IEEE International Conference on*, volume 4, pages 2448–2453. IEEE, 1994.

[3] Andrew G. Barto and Sridhar Mahadevan. Recent advances in hierarchical reinforcement learning. *Discrete Event Dynamic Systems*, 13(4):341–379, 2003.

[4] M. G. Bellemare, Y. Naddaf, J. Veness, and M. Bowling. The arcade learning environment: An evaluation platform for general agents. *Journal of Artificial Intelligence Research*, 47:253–279, 06 2013.

[5] Marc G Bellemare, Georg Ostrovski, Arthur Guez, Philip S Thomas, and Rémi Munos. Increasing the action gap: New operators for reinforcement learning. *arXiv preprint arXiv:1512.04860*, 2015.

[6] Matthew J. Hausknecht and Peter Stone. Deep Recurrent Q-Learning for Partially Observable MDPs. *CoRR*, abs/1507.06527, 2015.

[7] Milos Hauskrecht, Nicolas Meuleau, Leslie Pack Kaelbling, Thomas Dean, and Craig Boutilier. Hierarchical solution of Markov decision processes using macro-actions. In *Proceedings of the Fourteenth conference on Uncertainty in artificial intelligence*, pages 220–229. Morgan Kaufmann Publishers Inc., 1998.

[8] Tejas D Kulkarni, Karthik R Narasimhan, Ardavan Saeedi, and Joshua B Tenenbaum. Hierarchical deep reinforcement learning: Integrating temporal abstraction and intrinsic motivation. *arXiv preprint arXiv:1604.06057*, 2016.

[9] Volodymyr Mnih, Adria Puigdomenech Badia, Mehdi Mirza, Alex Graves, Timothy P Lillicrap, Tim Harley, David Silver, and Koray Kavukcuoglu. Asynchronous methods for deep reinforcement learning. *arXiv preprint arXiv:1602.01783*, 2016.

[10] Volodymyr Mnih, Koray Kavukcuoglu, David Silver, Alex Graves, Ioannis Antonoglou, Daan Wierstra, and Martin Riedmiller. Playing atari with deep reinforcement learning. *arXiv preprint arXiv:1312.5602*, 2013.

[11] Volodymyr Mnih, Koray Kavukcuoglu, David Silver, Andrei A. Rusu, Joel Veness, Marc G. Bellemare, Alex Graves, Martin Riedmiller, Andreas K. Fidjeland, Georg Ostrovski, Stig Petersen, Charles Beattie, Amir Sadik, Ioannis Antonoglou, Helen King, Dharshan Kumaran, Daan Wierstra, Shane Legg, and Demis Hassabis. Human-level control through deep reinforcement learning. *Nature*, 518(7540):529–533, February 2015.

[12] Arun Nair, Praveen Srinivasan, Sam Blackwell, Cagdas Alcicek, Rory Fearon, Alessandro De Maria, Vedavyas Panneershelvam, Mustafa Suleyman, Charles Beattie, Stig Petersen, et al. Massively parallel methods for deep reinforcement learning. *arXiv preprint arXiv:1507.04296*, 2015.

[13] Ian Osband, Charles Blundell, Alexander Pritzel, and Benjamin Van Roy. Deep exploration via bootstrapped dqn. *arXiv preprint arXiv:1602.04621*, 2016.

[14] Doina Precup. Temporal abstraction in reinforcement learning, 2000.

[15] Tom Schaul, John Quan, Ioannis Antonoglou, and David Silver. Prioritized experience replay. *arXiv preprint arXiv:1511.05952*, 2015.

[16] Richard S. Sutton, Doina Precup, and Satinder Singh. Between MDPs and semi-MDPs: A framework for temporal abstraction in reinforcement learning. *Artificial intelligence*, 112(1):181–211, 1999.

[17] Richard S Sutton, Doina Precup, and Satinder Singh. Between mdps and semi-mdps: A framework for temporal abstraction in reinforcement learning. *Artificial intelligence*, 112(1):181–211, 1999.

[18] T. Tieleman and G. Hinton. Lecture 6.5—RmsProp: Divide the gradient by a running average of its recent magnitude. COURSERA: Neural Networks for Machine Learning, 2012.

[19] Hado Van Hasselt, Arthur Guez, and David Silver. Deep reinforcement learning with double q-learning. *arXiv preprint arXiv:1509.06461*, 2015.

[20] Ziyu Wang, Nando de Freitas, and Marc Lanctot. Dueling network architectures for deep reinforcement learning. *arXiv preprint arXiv:1511.06581*, 2015.
[21] Christopher JCH Watkins and Peter Dayan. Q-learning. *Machine learning*, 8(3-4):279–292, 1992.